\documentclass[pdflatex,sn-mathphys-num]{sn-jnl}

\usepackage{graphicx}%
\usepackage{multirow}%
\usepackage{amsmath,amssymb,amsfonts}%
\usepackage{amsthm}%
\usepackage{mathrsfs}%
\usepackage[title]{appendix}%
\usepackage{xcolor}%
\usepackage{textcomp}%
\usepackage{manyfoot}%
\usepackage{booktabs}%
\usepackage{algorithm}%
\usepackage{algorithmicx}%
\usepackage{algpseudocode}%
\usepackage{listings}%
\usepackage{makecell}
\usepackage{adjustbox}
\usepackage[table]{xcolor}

\theoremstyle{thmstyleone}%
\theoremstyle{thmstyletwo}%

\theoremstyle{thmstylethree}%
\newtheorem{definition}{Definition}%

\begin{document}

\title[Article Title]{Autorelevance function and other feature relevance measures for univariate time series}


\author[1]{\fnm{Julian} \sur{Cardenas}}\email{julian.alonso.cardenas@upc.edu}

\author[2]{\fnm{Jamie} \sur{Arjona}}\email{jamie.arjona@upc.edu}

\author[3]{\fnm{Pedro} \sur{Delicado}}\email{pedro.delicado@upc.edu}

\affil*[1,2,3]{\orgdiv{Departament d'Estadística i Investigació Operativa}, \orgname{Universitat Politècnica de Catalunya - BarcelonaTech}, \orgaddress{\street{C/ Jordi Girona 1-3}, \city{Barcelona}, \postcode{08034}, \state{Catalunya}, \country{Spain}}}




\abstract{We propose a model agnostic methodology to measure lag relevance in machine learning forecasting models applied to univariate time series. Particularly, we are working in the context of time series using the frameworks of Ghost variables and Shapley values, together with additive importance measures, to introduce the auto-relevance and partial auto-relevance functions as the lag importance values. Additionally, we propose a novel method to replace absent features in coalition based methods with a one step forecast from the same model. We evaluate these proposals under different simulations and real data cases. This combined framework perspective is particularly suitable for time series. In addition, to show our discoveries we use a pull of models from the seasonal ARMA family and recurrent neural networks. We found that the calculated relevance measures successfully demonstrate the expected lag structure in almost all cases.}




\maketitle

\section{Introduction}\label{sec1}

The increasing use of machine learning models for prediction has intensified the need for transparency and interpretability in artificial intelligence. Although models such as neural networks and tree-based ensembles often outperform classical statistical approaches in predictive accuracy, they usually provide limited information about how the input variables influence the prediction. This lack of interpretability has motivated the rapid development of interpretable machine learning (IML) and explainable artificial intelligence (XAI), being one of the main objectives the quantification of the global relevance of each input variable in a predictive model \cite{molnar2025,biecek2021}.

This problem is especially important in time series forecasting because the predictors are typically lagged observations of the same series, and understanding the relevance of each lag is often directly connected to the interpretation of the underlying temporal structure. While classical tools such as the autocorrelation function (ACF) and partial autocorrelation function (PACF) provide valuable guidance in linear settings, they are not designed to explain the behavior of modern black-box forecasting models. Consequently, there is a growing need for model-agnostic relevance measures that can identify the contribution of lagged inputs in both linear and nonlinear forecasting frameworks.

This paper addresses that problem by proposing relevance measures for lagged variables in time series forecasting models. Our focus is on methodologies that quantify how each lag contributes to the predictive performance of a trained model, with particular attention to forecasting settings where the inputs consist exclusively of lagged values of the series. The proposed framework is designed to be model-agnostic and can therefore be applied to both classical statistical models and black-box predictors such as neural networks.

A key component of the proposed methodology is the treatment of features that are considered absent under a family of methods called \textit{coalition-based relevance methods}. These coalition-based methods divided the features into present or absent in the forecasting task with the goal of evaluate the relevance when they are in present or not in subsets of features. These subsets are called \textit{coalitions}. In particular, we consider imputation strategies that preserve the temporal structure of the series, including a model-driven one-step-ahead forecasting scheme. This idea allows the relevance analysis to remain aligned with the dynamics of the forecasting model itself and provides a natural way to define coalitions in sequential settings.

The remainder of the paper is organized as follows. Section~\ref{sec:related_work} reviews the main literature on global feature importance, Shapley-based relevance methods, and explainability for time series. Section~\ref{sec:methodology} introduces the proposed relevance measures and the associated imputation schemes. Section~\ref{sec2} presents the empirical evaluation based on simulated and real datasets. Finally, Section~\ref{sec:conclusion} concludes with a discussion of the main findings and possible directions for future research.


\section{Related Work}
\label{sec:related_work}

A large part of the literature on global feature importance is based on perturbation or removal strategies. Since the introduction of permutation variable importance  \cite{breiman2001random}, many approaches have quantified the relevance of a feature by evaluating the deterioration in predictive performance caused by its omission or perturbation. A representative example is the Leave-One-Covariate-Out (LOCO) methodology \cite{lei2018loco}, which measures the predictive impact of removing one variable at a time. Although intuitive, these methods may become unstable under correlated predictors and computationally demanding because model refitting is required, which has motivated more efficient variants like \textit{ghost variables} method \cite{delicado2023understanding}.

A more principled alternative is based on Shapley values from cooperative game theory \cite{shapley1953value}. In this framework, the input variables are interpreted as players in a game, and the relevance of each variable is defined through its average marginal contribution across all feature coalitions. In machine learning, \cite{lundberg2017unified} popularized this idea through SHAP, focused on local interpretability. Additionally, showing that Shapley values provide a unique additive attribution rule under a set of natural axioms. Global relevance can then be obtained by aggregating local Shapley-based contributions, which connects naturally with the broader framework of additive importance measures (AIM) introduced by \cite{AIM2022}.

An important issue in Shapley-based explanations is how to impute absent features. In a forecasting problem, generating coalitions with absent features implies that there should be an imputation method to replace these features. Different imputation schemes may lead to substantially different interpretations of relevance. For example, \cite{Janzing2020Feature} emphasize that alternative imputations are associated with different explanatory and causal meanings, while \cite{Williamson2020ShapleyConditional} study the role of conditional expectations in nonparametric Shapley-based inference. In the context of global importance, \cite{AIM2022} provide a theoretical justification for loss-based additive relevance measures, particularly under mean squared error (MSE), which is especially attractive in forecasting problems.

For time series data, several recent works have adapted Shapley-based ideas to account for temporal dependence. TimeSHAP \cite{bento2021timeshap} and WindowSHAP \cite{nayebi2023windowshap} are local interpretability methods and TsSHAP \cite{raykar2023tsshap} generalizes to global interpretability, but of all them propose variants of SHAP that aggregate or mask temporal information to identify influential time steps. Related work has also emphasized the importance of defining coalitions and imputations in ways that respect temporal order \cite{villani2022feature}, and SHAP-based methods have already been applied in practical forecasting contexts such as sales and demand prediction \cite{arboleda2023interpreting}.

Despite these advances, the treatment of absent features in time-series explainability remains largely heuristic. Existing approaches often rely on generic imputations such as zero values, unconditional means, or marginal sampling. In contrast, our work considers imputation strategies that are explicitly adapted to forecasting settings, including a model-based one-step-ahead imputation mechanism that preserves temporal coherence. In addition, by combining these imputations with relevance measures defined through predictive loss, the proposed framework connects the practical needs of time-series explainability with the theoretical foundations provided by additive importance measures.


\section{Methods}
\label{sec:methodology}

Our objective is to define global lag-relevance measures for univariate time series given a trained one-step-ahead forecasting model. In particular, we seek \textit{model-agnostic} relevance measures, that is, measures that can be computed regardless of the internal structure of the forecasting model. Throughout this work, we assume that the prediction function depends only on lagged values of the same series. Although the exposition is restricted to one-step-ahead forecasting, the methodology can be extended naturally to the $k$-step-ahead case.

\subsection{Lag relevance based on ghost variables}

Following \cite{delicado2023understanding}, we adapt the ghost variables methodology to the univariate time series forecasting setting. Let $\{y_t\}_{t=1}^T$ denote a univariate time series of length $T$. We consider one-step-ahead prediction of $y_t$ based on its last $p$ lags through a fitted prediction function $\hat{f}$, estimated using a training sample $S_1 =\{1,\ldots,n_1\}$ of size $n_1$. Let $S_2 = \{n_1+1,\ldots,T\}$ be a disjoint test sample, over which lag relevance is evaluated. For each $t \in S_2$, the standard one-step-ahead prediction is

\begin{equation}
\hat{y}_t = \hat{f}(y_{t-1},\ldots,y_{t-p}).
\label{eq:pred-original-short}
\end{equation}

To evaluate the relevance of lag $h$, we replace $y_{t-h}$ by a ghost version and define the alternative prediction
\begin{equation}
\hat{y}^{(g)}_{t,h}
=
f(\ldots,y_{t-h+1},\hat{y}^{*}_{t-h},y_{t-h-1},\ldots).
\label{eq:pred-ghost-short2}
\end{equation}
Here, $\hat{y}^{*}_{t-h}$, known as \textit{ghost value}, denotes an imputed value of the $h$th lag. This ghost value is obtained by predicting $y_{t-h}$ from the remaining $p-1$ lags. In this work, the auxiliary imputation model is a multiple linear regression of the form

\small{
\begin{equation}
\hat{y}^{*}_{t-h}
=
\hat{\beta}_0
+
\sum_{i \in M \setminus \{h\}}
\hat{\beta}_i\, y_{t-i},
\label{eq:ghost-linear-compact}
\end{equation}}
where $M=\{1,\ldots,p\}$. The relevance of lag $h$ is then quantified by comparing the original predictions in \eqref{eq:pred-original-short} with the ghost-based predictions in \eqref{eq:pred-ghost-short2}, with all of these elements we can introduce the following definition. 

\begin{definition}
For the collection of values $h=1,\ldots,p$, we define the ghost-variables autorelevance function with MSPD contributions as:
\small{
\begin{equation}
ARF_h^{\scriptscriptstyle \mathrm{MSPD}}
=
\tfrac{1}{T-n_1}
\sum_{t=n_1+1}^{T}
\left(\hat{y}_t-\hat{y}^{gh}_{t,h}\right)^2.
\label{eq:arf-gh-mspd-compact}
\end{equation}
}   
\end{definition}


As discussed in \cite{delicado2023understanding}, MSPD-type measures are closely related to changes in mean squared error (MSE), which is the most commonly used loss in forecasting. Defining $d^{(h)}_t = (y_t-\hat{y}^{gh}_{t,h})^2 - (y_t-\hat{y}_t)^2$, this $d^{(h)}_t$ motivates an alternative definition.

\begin{definition}
For the collection of values $h=1,\ldots,p$, the quantities 

\begin{equation}
ARF_h^{\scriptscriptstyle \mathrm{MSE}}
=
\tfrac{1}{T-n_1}
\sum_{t=n_1+1}^{T}
d_t^{(h)},
\label{eq:arf-gh-mse-compact}
\end{equation}

will be called the ghost-variables autorelevance function with MSE contributions.
\end{definition}

\subsection{Lag relevance based on Shapley values}

We now introduce a Shapley-based relevance measure adapted to forecasting settings. Instead of defining marginal contributions through changes in the model output, we evaluate them through changes in prediction error. This modification is motivated by the additive importance measures framework of \cite{AIM2022} and by the central role of MSE in time series forecasting.

Let $M=\{1,\ldots,p\}$ denote the set of available lags, and let $S \subseteq M$ be a coalition of lags. For each coalition $S$, let $\hat{f}_t(S)$ denote the prediction obtained when the lags in $S$ are retained and the remaining lags are imputed. We also define the baseline prediction
\begin{equation}
\hat{f}_t(\bar{y})
=
\hat{f}_t(\bar{y},\ldots,\bar{y}),
\label{eq:baseline-pred-compact}
\end{equation}
where $\bar{y}$ denotes the sample mean computed on the training set.

Considering $\Delta_t^{\scriptscriptstyle \mathrm{MSE}}(S) = (\hat{f}_t(\bar{y})-y_t)^2 - (\hat{f}_t(S)-y_t)^2$. The value function based on MSE improvement is
\begin{equation}
v_{\scriptscriptstyle \mathrm{MSE}}(S)
=
\tfrac{1}{T-n_1}
\sum_{t=n_1+1}^{T}
\Delta_t^{\scriptscriptstyle \mathrm{MSE}}(S).
\label{eq:value-mse-compact}
\end{equation}

Thus, $v_{\scriptscriptstyle \mathrm{MSE}}(S)$ measures the improvement in prediction error achieved by coalition $S$ with respect to the baseline.

To simplify notation, we define for $S \subseteq M \setminus \{h\}$, the Shapley weight
\begin{equation}
\omega(S,h)
=
\frac{|S|!(p-|S|-1)!}{p!}.
\label{eq:omega-def}
\end{equation}

The Shapley relevance of lag $h$ is then
\begin{equation}
\begin{aligned}
\phi_h(v_{\scriptscriptstyle \mathrm{MSE}})
&=
\sum_{S \subseteq M\setminus\{h\}}
\omega(S,h) \cdot
\bigl(v_{\scriptscriptstyle \mathrm{MSE}}(S\cup\{h\})-v_{\scriptscriptstyle \mathrm{MSE}}(S)\bigr).
\end{aligned}
\label{eq:shapley-mse-compact2}
\end{equation}

For interpretability, we also define the incremental loss reduction

\begin{equation}
\begin{aligned}
\Delta_{\scriptscriptstyle \mathrm{MSE}}(S,h)
&=
\tfrac{1}{T-n_1}
\sum_{t=n_1+1}^{T}
\bigl((y_t-\hat{f}_t(S))^2 
-(y_t-\hat{f}_t(S\cup\{h\}))^2\bigr).
\end{aligned}
\label{eq:delta-mse}
\end{equation}

Note that when $S=\{1,\cdots,p\} \setminus \{h\}$, this calculation coincide with the definition or ARF in \eqref{eq:arf-gh-mse-compact}. Returning to the central point, integrating Equations \eqref{eq:shapley-mse-compact2} and \eqref{eq:delta-mse}, we obtain

\begin{equation}
\phi_h(v_{\scriptscriptstyle \mathrm{MSE}})
=
\sum_{S \subseteq M\setminus\{h\}}
\omega(S,h)\,
\Delta_{\scriptscriptstyle \mathrm{MSE}}(S,h).
\label{eq:shapley-mse-delta}
\end{equation}
Using the Shapley efficiency property, we have
\begin{equation}
v_{MSE}(M)
=
\sum_{h=1}^{p}\phi_h(v_{\scriptscriptstyle \mathrm{MSE}}).
\label{eq:efficiency-mse-compact}
\end{equation}
Therefore, we can define the standardized relevance measure as follows.

\begin{definition}
For $h=1,\ldots,p$, the values $\tilde{\phi}_h(v_{\scriptscriptstyle \mathrm{MSE}})$, represent the \textit{partial autorelevance function} (PARF) with MSE contributions as:

\begin{align}
& \tilde{\phi}_h(v_{\scriptscriptstyle \mathrm{MSE}})
=\frac{\phi_h(v_{\scriptscriptstyle \mathrm{MSE}})}{v_{\scriptscriptstyle \mathrm{MSE}}(M)},
& \sum_{h=1}^{p}\tilde{\phi}_h(v_{\scriptscriptstyle \mathrm{MSE}})=1.
\label{eq:parf-mse-compact2}
\end{align}

\end{definition}

The term \textit{partial} is used by analogy with the partial autocorrelation function. In the ghost-variables approach, relevance is assessed by replacing one lag at a time while keeping all others fixed. In contrast, the Shapley framework evaluates the contribution of each lag across all possible subsets, thereby capturing both direct and indirect effects through the coalition structure.

Similarly to ARF, we define $\delta_t(S,h)=\hat f_t(S\cup\{h\})-\hat f_t(S)$ to obtain a Shapley-type relevance measure based on squared prediction differences, analogous to the ghost-variable MSPD criterion,
\begin{equation}
\Delta^{\mathrm{MSPD}}(S,h)
=
\tfrac{1}{T-n_1}
\sum_{t=n_1+1}^{T}
\delta_t(S,h)^2.
\label{eq:delta-mspd-compact}
\end{equation}
and
\begin{equation}
\phi_h(v_{\scriptscriptstyle \mathrm{MSPD}})
=
\sum_{S \subseteq M\setminus\{h\}}
\omega(S,h)\,
\Delta_{\scriptscriptstyle \mathrm{MSPD}}(S,h).
\label{eq:shapley-mspd-compact2}
\end{equation}
We need to mention that MSPD is valid as a relevance criterion. But, this quantity should not be interpreted as a classical cooperative-game value function in the strict sense, because, the value assigned to a coalition is not an intrinsic payoff produced by that coalition, but a discrepancy between two predictive scenarios: the original model prediction and the prediction under a modified input, this not guarantee the monotonicity required to be a value function. However, we can define the last relevance measure with the MSPD contribution value function.
\begin{definition}
For $h=1,\ldots,p$, the values $\tilde{\phi}_h(v_{\scriptscriptstyle \mathrm{MSPD}})$, represent the \textit{partial autorelevance function} (PARF) with MSPD contributions as:
\begin{equation}
\tilde{\phi}_h(v_{\scriptscriptstyle \mathrm{MSPD}})
=
\frac{\phi_h(v_{\scriptscriptstyle \mathrm{MSPD}})}
     {v_{\scriptscriptstyle \mathrm{MSPD}}(M)-v_{\scriptscriptstyle \mathrm{MSPD}}(\emptyset)}.
\label{eq:parf-mspd-compact2}
\end{equation}
with
\begin{equation}
v_{\scriptscriptstyle \mathrm{MSPD}}(M)-v(\emptyset)
=
\tfrac{1}{T-n_1}
\sum_{t=n_1+1}^{T}
\bigl[
\hat{f}_t-\hat{f}_t(\bar{y})
\bigr]^2.
\label{eq:mspd-normalizer-compact2}
\end{equation}    
\end{definition}

Using the same argument as in the $MSE$ value function case, under the normalization in Equation \eqref{eq:parf-mspd-compact2}, we can affirm that,
\begin{equation}
\sum_{h=1}^{p}\tilde{\phi}_h(v_{\scriptscriptstyle \mathrm{MSPD}}) \approx 1.
\label{eq:mspd-approx-one-compact}
\end{equation}

This standardization allows direct comparison between the relevance profiles obtained from ghost variables and those derived from the Shapley-based construction. Finally, unlike standard SHAP procedures, which aggregate local contributions across observations, the proposed formulation computes relevance directly at the global level through averages over the test sample. This substantially reduces the computational burden due to, in local SHAP-based methods, all permutations of the predictors must be examined for each point in the test sample, while with our approach we only need to examine the permutations of the predictors once. In practice, additional gains can be achieved by approximating the Shapley sums through random permutations when the number of lags is large.


\section{Results}\label{sec2}

In this section, we evaluate the proposed relevance measures under the different contribution definitions and imputation strategies using both simulated and real time series data. The simulated examples include linear, nonlinear, and seasonal structures, while the real-data applications are intended to illustrate the practical behavior of the methodology. Our main goal is to assess whether the proposed measures are able to recover meaningful lag-importance patterns, particularly in black-box forecasting models.\\

Throughout the experiments, all forecasting models are restricted to use only lagged values of the series as inputs, so that the resulting relevance scores can be interpreted directly as measures of lag importance. In addition to black-box models, we also consider classical ARMA-type models, which provide a natural benchmark in well-understood settings. These ARMA models are selected automatically under automatic procedures based on \textit{auto.arima} function implemented in the \textit{pmdarima} Python package \cite{pmdarima}. Additionally, for these models, Shapley-based relevance measures are computed using a surrogate linear model that approximates the one-step-ahead forecasts of the fitted ARMA predictor. This avoids repeated refitting within each coalition and yields a stable approximation of the predictive mapping.\\

For the black-box setting, we focus mainly on neural network forecasting models with different architectures. Before computing the relevance measures, all models were carefully tuned to ensure satisfactory predictive performance. Hyperparameter selection was carried out using a validation set, so that model selection and final evaluation remained properly separated. After tuning, the relevance measures and forecasting metrics were computed on the full test set. The final hyperparameter values, architectural details, and out-of-sample predictive results are reported into Appendix~B and can be complemented in GitHub repository created for this work\footnote{\href{Autorelevance}{https://github.com/alojulian15/Autorelevance}}.\\

In the main text, we report the auto-relevance function (ARF) and partial auto-relevance function (PARF) obtained with the one-step-ahead imputation method, since this strategy provided the most stable and informative results across the considered scenarios. The corresponding results based on linear imputation are included in Appendix~A.

\subsection{Linear simulations}

To generate linear time series data, we simulate observations from models belonging to the ARMA family, assuming a noise standard deviation of $\sigma = 0.5$ in each scenario. For every case, a time series of $1000$ observations is generated and subsequently divided into training and test sets. The training set consists of the first $80\%$ of the observations, while the remaining $20\%$ are reserved for the test set.\\

As first example, we consider an AR$(3)$ model with autoregressive parameters $\phi_1 = 0.4$, $\phi_2 = 0$, and $\phi_3 = -0.5$. This specification is especially interesting because the second lag does not contribute directly to the data-generating process. Consequently, this setting allows us to examine whether the proposed relevance measures are capable of identifying the negligible contribution of the second lag.\\

\begin{figure}[h!]
\centering
\includegraphics[width=0.8\linewidth]{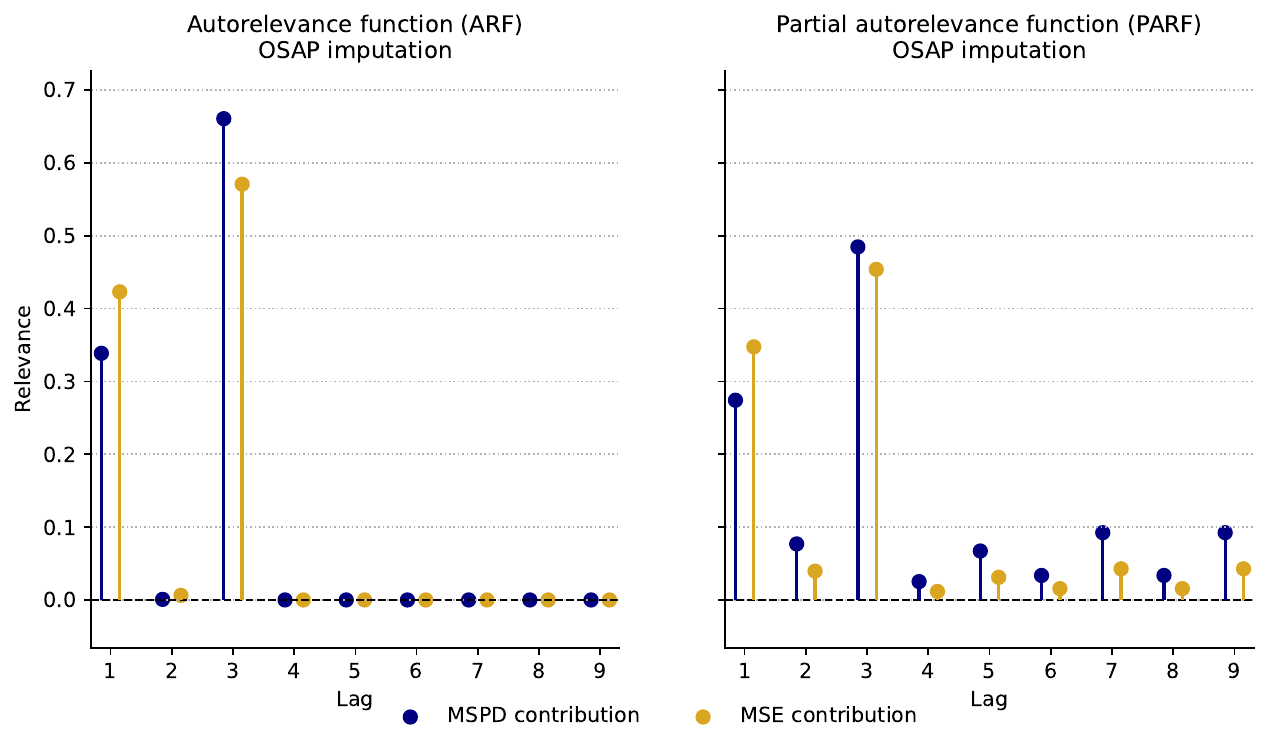}
\caption{Autorelevance and partial autorelevance functions - ARMA fitted model - AR$(3)$ series}
\label{fig:siete}
\end{figure}

We compute the ARF and PARF using an automatically selected ARMA model as the prediction function. The resulting plots are shown in Figure~\ref{fig:siete}. As expected, the relevances associated with lags $1$ and $3$ are clearly different from zero in the ARF. More importantly, the relevance of the second lag remains close to zero, which is consistent with the underlying data-generating mechanism. A similar pattern can be observed in the PARF: although some additional lags exhibit small non-zero values, the dominant relevance values are associated with lags $1$ and $3$. These results indicate that the proposed methods are able to recover the main lag structure of the series when the fitted model corresponds closely to the process that generated the data.\\

\begin{figure}[h!]
\centering
\includegraphics[width=0.8\linewidth]{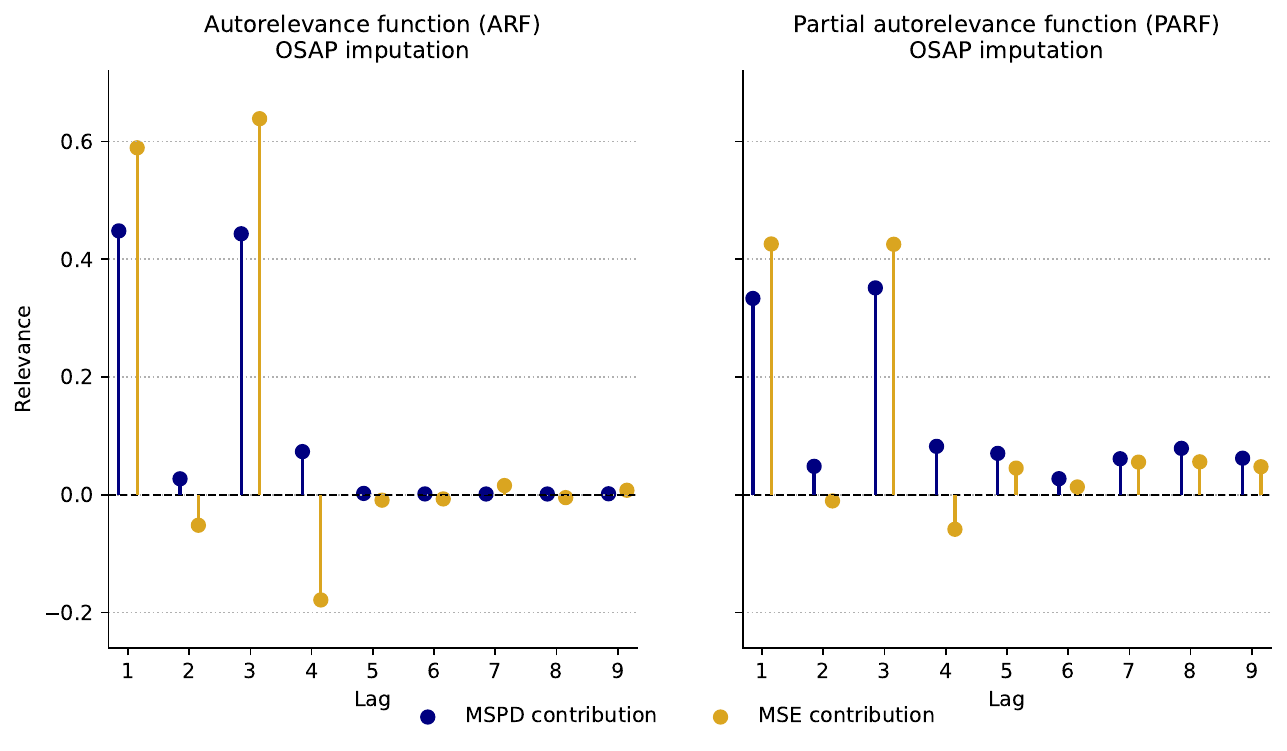}
\caption{Autorelevance and partial autorelevance functions - NN fitted model - AR$(3)$ series}
\label{fig:nueve}
\end{figure}

Using the same simulated dataset, we also fit a neural network forecasting model, specifically an LSTM architecture with two hidden layers. The corresponding ARF and PARF are displayed in Figure~\ref{fig:nueve}. The resulting plots again show that the relevance values associated with lags $1$ and $3$ are substantially larger than those of the remaining lags. Overall, these results suggest that the proposed auto-relevance measures are capable of identifying the key predictive lags even when the forecasting model is a black-box neural network. In all cases, the two largest relevance values correspond to lags $1$ and $3$, which is fully consistent with the structure imposed in the simulation.

\begin{figure}[h!]
\centering
\includegraphics[width=0.6\linewidth]{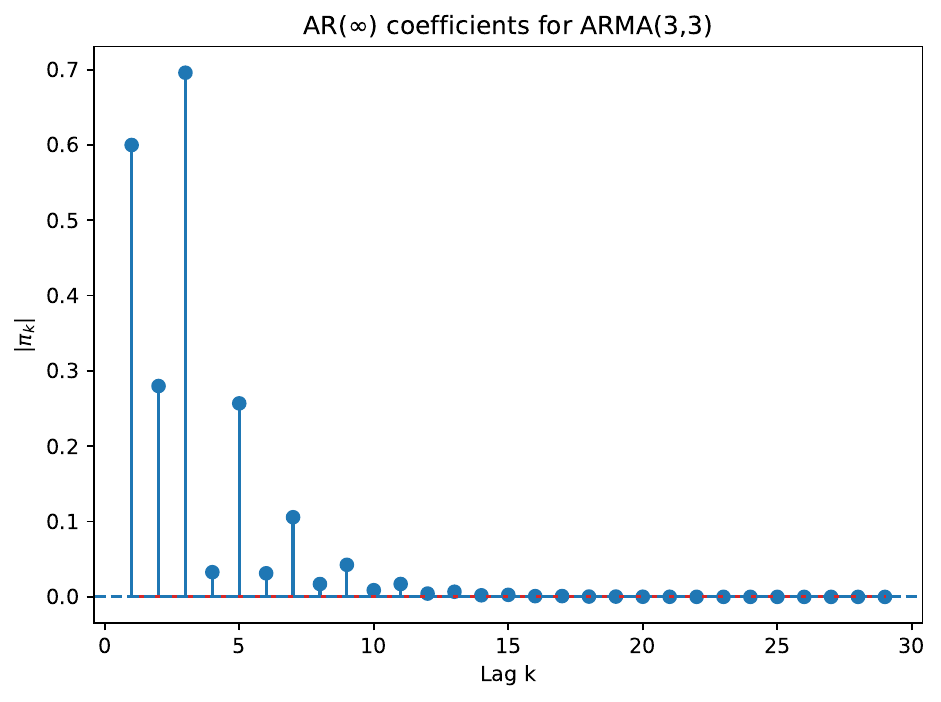}
\caption{Absolute value of coefficients from AR$(\infty)$ representations for ARMA simulated model}
\label{fig:InfAR_ex05}
\end{figure}

Now, we consider data generated from an ARMA$(3,3)$ model. The autoregressive parameters are $\phi_1 = 0.4$, $\phi_2 = 0$, and $\phi_3 = -0.5$, which coincide with those used in the previous AR example. In addition, the moving average component is specified with parameters $\theta_1 = 0.2$, $\theta_2 = 0.4$, and $\theta_3 = 0.1$. To have a basis of comparison, we show the infinite AR representation of this simulated process into Figure~\ref{fig:InfAR_ex05}. As observed in the last example, the behavior of the proposed relevance measures provides clear indications of the most influential lags in the series.\\

\begin{figure}[h!]
\centering
\includegraphics[width=0.8\linewidth]{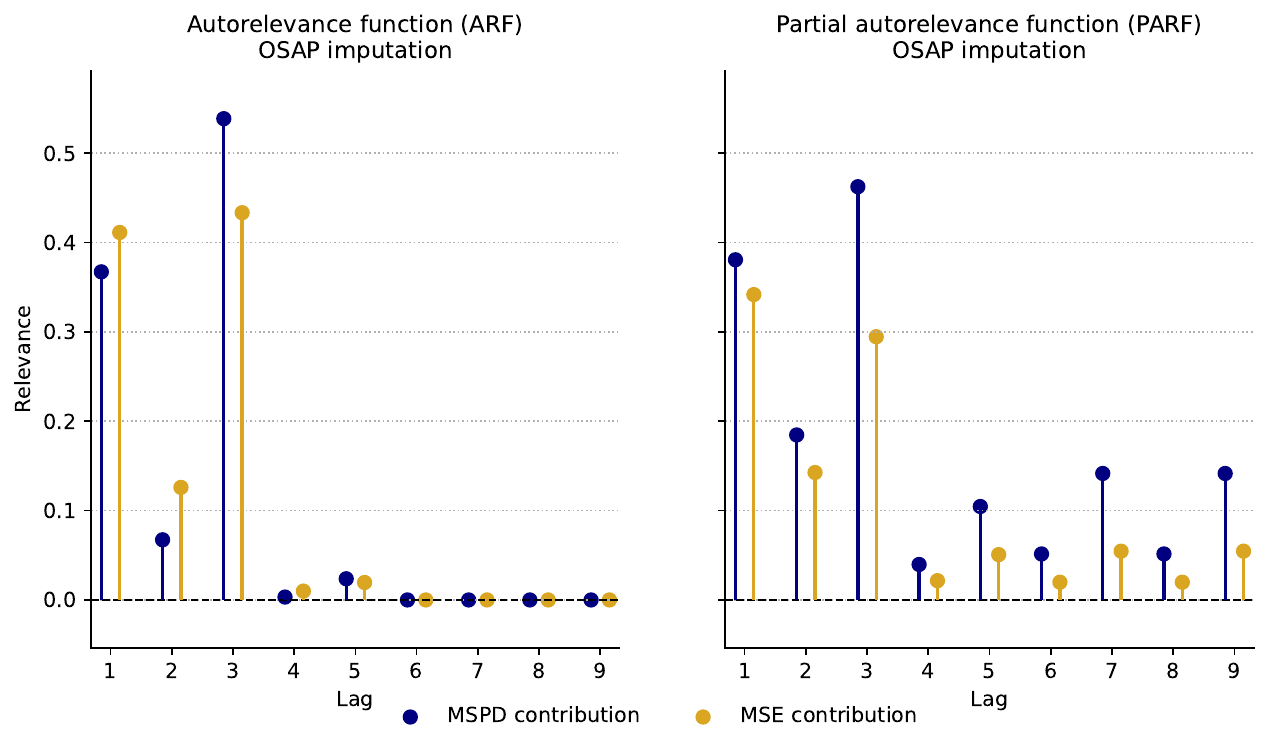}
\caption{Autorelevance and partial autorelevance functions - ARMA fitted model - ARMA$(3,3)$ series}
\label{fig:catorce}
\end{figure}

Figure~\ref{fig:catorce} presents the ARF and PARF obtained when the prediction model is an automatically selected ARMA specification. In this case, lags $1$ and $3$ appear consistently as the most relevant across all configurations, although the second lag also exhibits a small but noticeable relevance value. When examining the fitted automatic ARMA model, we observe that the selected specification has an autoregressive order of $5$. Despite this discrepancy between the fitted model and the true data-generating process, the relevance measures still capture the main structure of the underlying series.\\

\begin{figure}[h!]
\centering
\includegraphics[width=0.8\linewidth]{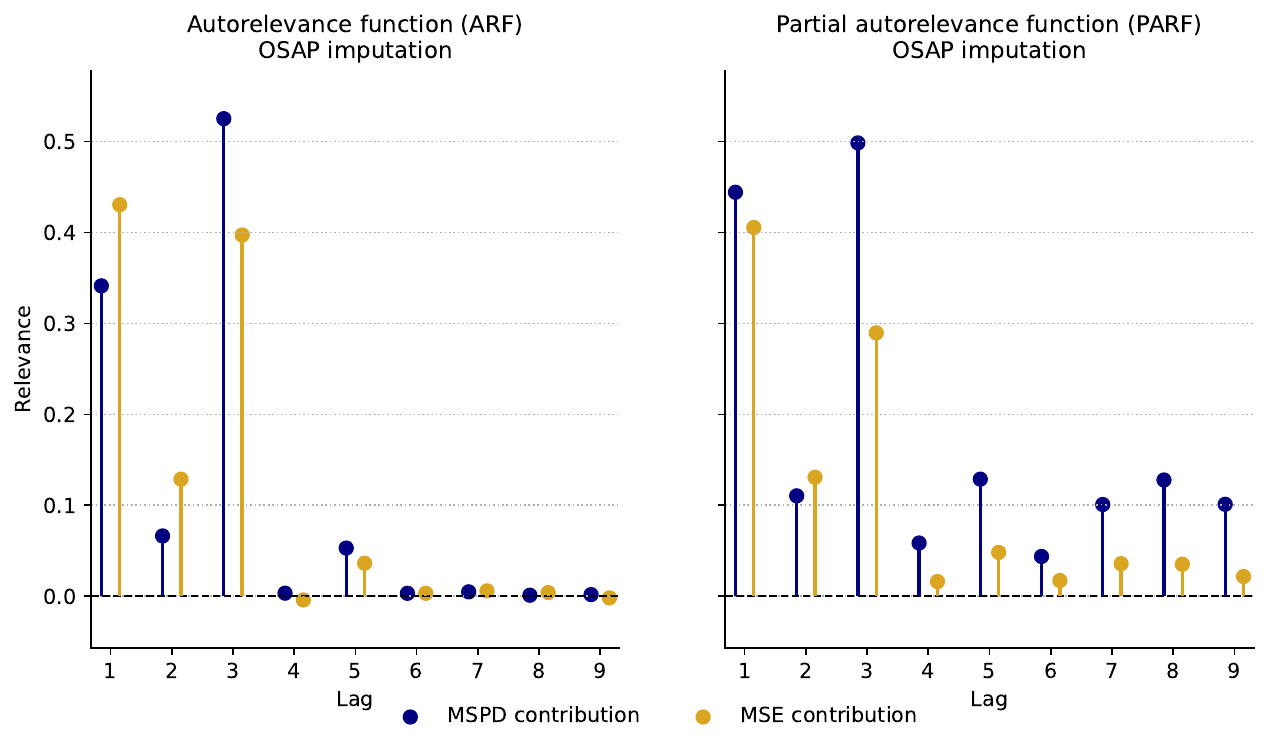}
\caption{Autorelevance and partial autorelevance functions - NN fitted model - ARMA$(3,3)$ series}
\label{fig:quince}
\end{figure}

When the forecasting model is a neural network (see Figure~\ref{fig:quince}), the relevance measures again highlight lags $1$ and $3$ as the most important predictors. In this scenario, the MSE-based relevance appears to follow more closely the lag structure, which is illustrated by the infinite AR representation displayed into Figure~\ref{fig:InfAR_ex05}.

\subsection{Non-linear time series}

To extend the analysis beyond ARMA and seasonal ARMA settings, we also consider simulated nonlinear time series. The goal is to evaluate whether the proposed relevance measures remain informative when the underlying data-generating process is nonlinear but still depends exclusively on lagged values of the series. As in the previous experiments, the simulated dataset has $n=1000$ and is divided into training and test sets using an $80\%$–$20\%$ split.

The simulated series is generated as follows. Let $\{y_t\}$ denote the simulated time series. The process follows a nonlinear autoregressive model whose parameters depend on a regime determined by the seasonal lag $y_{t-12}$. The regime indicator is defined as

\[
R_t =
\begin{cases}
\text{Low regime}, & y_{t-12} \le 0, \\
\text{High regime}, & y_{t-12} > 0 .
\end{cases}
\]

Conditional on the regime, the latent variable $z_t$ is defined as

\[
z_t =  v^{(R_t)} y_{t-1}y_{t-12}
+ 0.15 +
\sum_{i \in \{1,12,13\}} w_i^{(R_t)} \tanh(y_{t-i})
,
\]

where the coefficients differ across regimes are: In the \textbf{low regime} $w_1=0.9, w_{12}=-0.3, w_{13}=0.15, v=0.15$, while in the \textbf{high regime} $w_1=0.3, w_{12}=-0.9, w_{13}=0.45, v=0.60$. The observed series is obtained through a nonlinear link function with additive noise,

\[
y_t = \tanh(z_t) + \varepsilon_t,
\qquad
\varepsilon_t \sim N(0,0.18^2).
\]

Finally, simulated values are bounded to the interval $[-3,3]$ using the  clipping operator defined as $y_t=\min(3,\max(-3,y_t))$. The resulting process combines threshold regime switching and nonlinear autoregressive effects.\\

\begin{figure}[h!]
\centering
\includegraphics[width=0.8\linewidth]{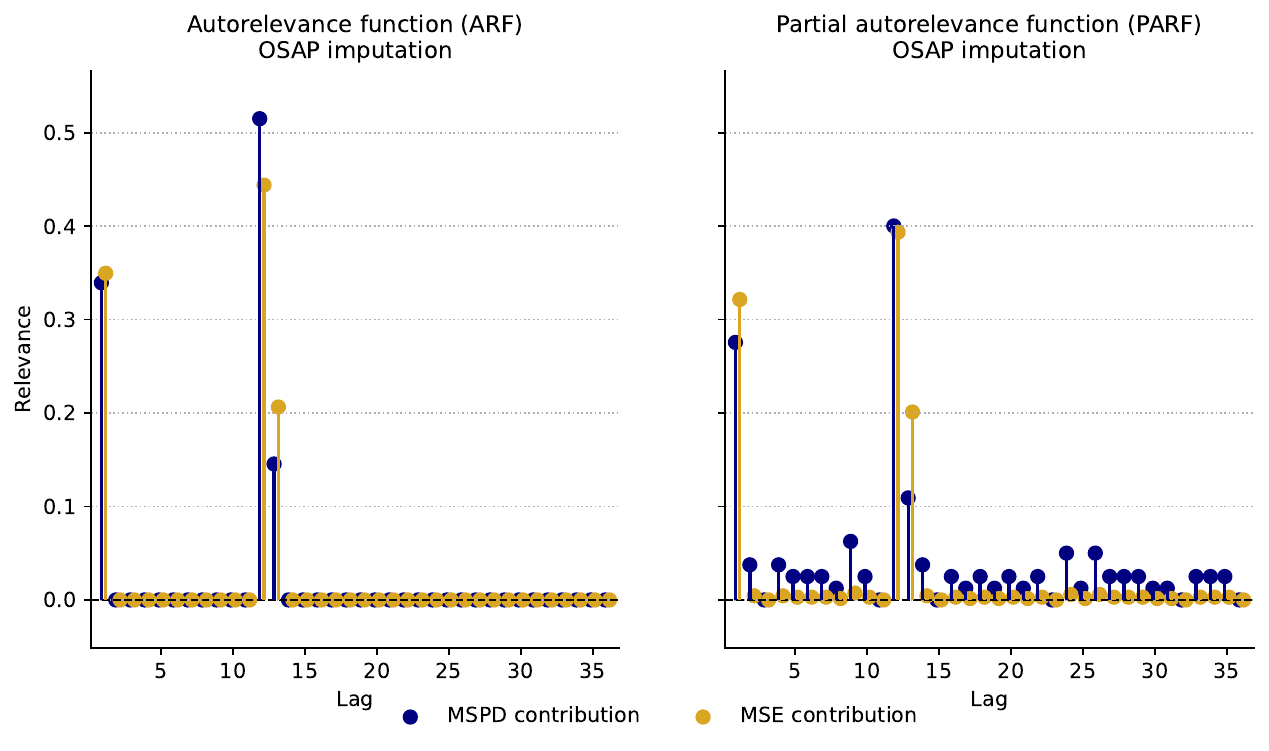}
\caption{Autorelevance and partial autorelevance functions - ARMA fitted model - Non linear time series}
\label{fig:nnl2arfar}
\end{figure}

When we analyze the relevance functions obtained using an automatically selected ARMA model as the prediction function (Figure~\ref{fig:nnl2arfar}), we observe that the relevance measures assign non-negligible values primarily to the lags that were expected to be important in the data-generating process. This indicates that, even when a linear model is used as the fitted predictor for a nonlinear process, the proposed relevance measures are still able to recover meaningful lag structures.

\begin{figure}[h!]
\centering
\includegraphics[width=0.8\linewidth]{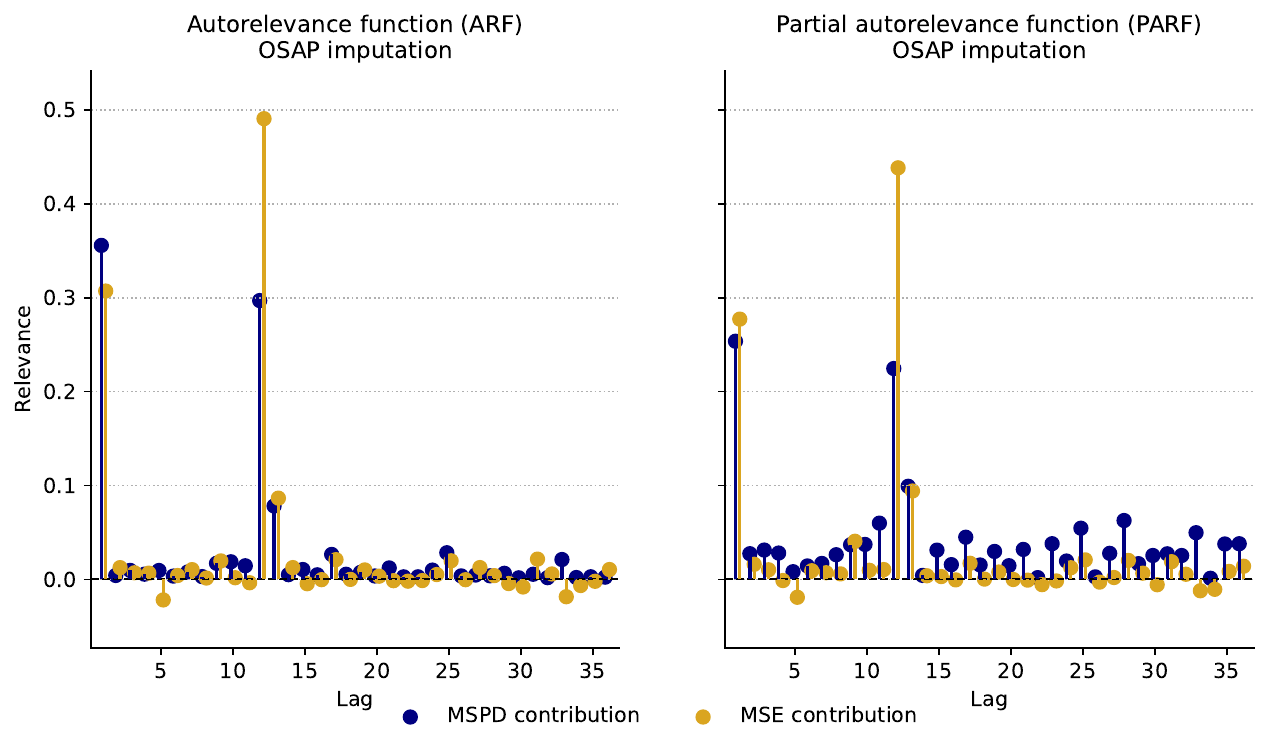}
\caption{Autorelevance and partial autorelevance functions - NN fitted model - Non linear time series}
\label{fig:nnl2arflstm}
\end{figure}

Figure~\ref{fig:nnl2arflstm} reports the results obtained when the forecasting model is a neural network. In this case, the relevance measures clearly identify lags $1$, $12$, and $13$ as the most important across all configurations. Moreover, when the relevance is computed using the MSE-based contributions, the measures capture this pattern more clearly, providing a sharper identification of the dominant lag structure of the simulated process.

\subsection{Application in real data}

The Monthly Airline Passengers dataset is a classic time series that records the total number of international airline passengers, in thousands, from January 1949 to December 1960. It contains 144 monthly observations and is widely used as a benchmark for forecasting and time series modeling. The typical transformations applied to this series are the logarithm, the first difference and a seasonal difference of order $12$. We fit the models for transformed data.

\begin{figure}[H]
\centering
\includegraphics[width=0.8\linewidth]{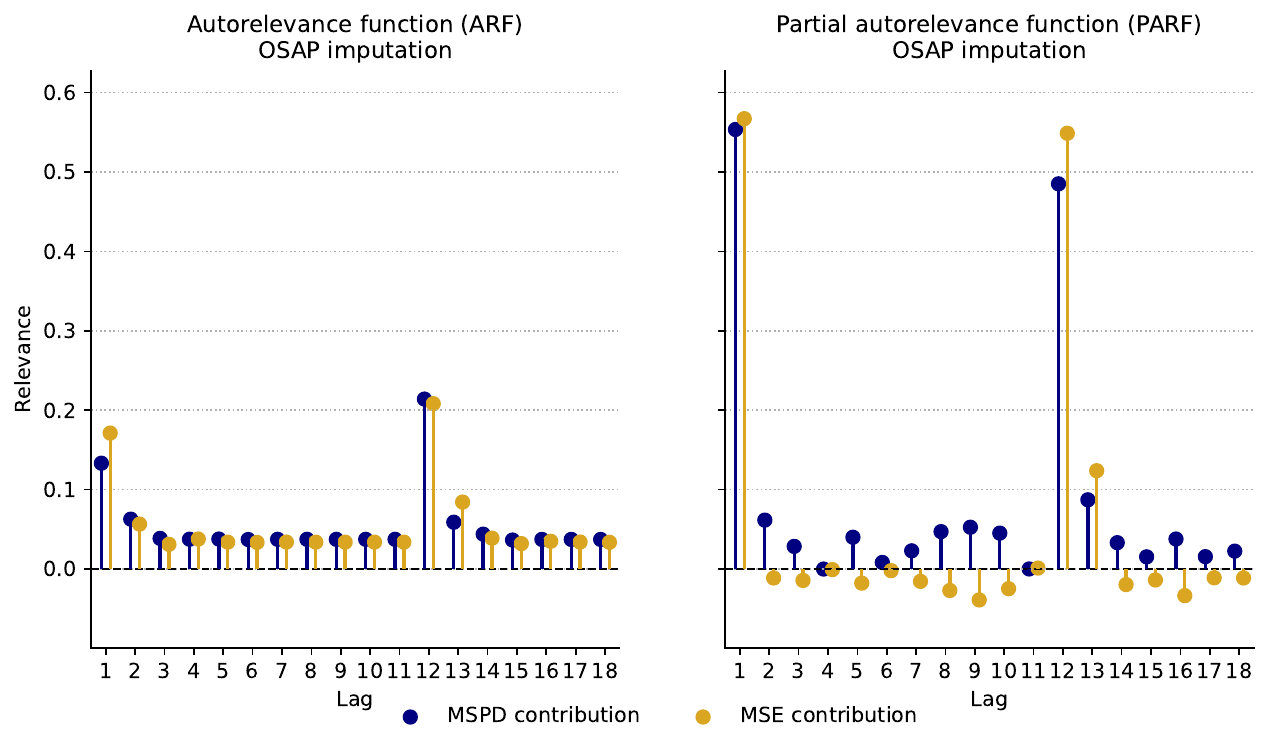}
\caption{Autorelevance and partial autorelevance functions - ARMA fitted model - Airline passengers series}
\label{fig:airline1}
\end{figure}

Under these transformations, the seasonal ARMA model selected for the series was $\mathrm{SARIMAX}(0,0,1) \times (0,0,1)_{12}$, which corresponds to the classical specification commonly used to model this dataset. The associated ARF and PARF are presented in Figure~\ref{fig:airline1}. Across all contribution definitions, the relevance measures consistently highlight lags $1$, $12$, and $13$ as the most important. In particular, the PARF identifies these lags more clearly than the ARF. In addition, when the MSPD-based contribution is used, the relevance measures assign a moderate value to lag $9$.

\begin{figure}[H]
\centering
\includegraphics[width=0.8\linewidth]{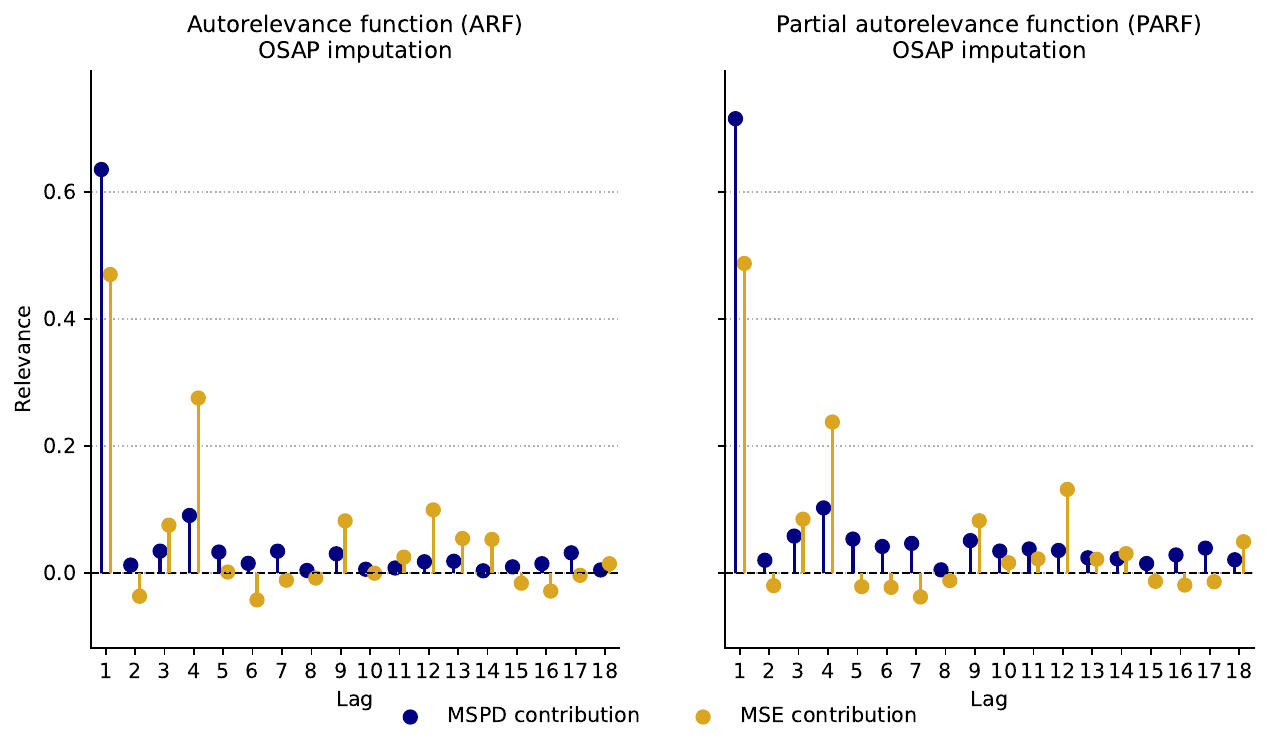}
\caption{ARF and Shapley based relevance plots - NN fitted model - Airline passengers series}
\label{fig:airline2}
\end{figure}

As in the previous experiments, we also fit a neural network model to the transformed series. The resulting relevance measures are shown in Figure~\ref{fig:airline2}. In this case, when the one-step-ahead imputation method is used, the relevance patterns resemble the behavior observed in this data, highlighting lags $1$, $9$, $12$, and $13$ as the most relevant. This produces similar signals between the two modeling approaches but the results derived from ARIMA model are more consistent with the expected behavior.

Nevertheless, this dataset has been extensively studied under the SARIMA specification described above, which is known to provide an excellent fit to the series. In contrast, the neural network model does not achieve the same level of forecasting performance. Taking this into account, the proposed relevance measures still appear to capture the fundamental lag structure of the series, regardless of the forecasting model employed.

\section{Conclusions}
\label{sec:conclusion}

The empirical results suggest that the ARF and PARF constitute effective tools for explaining forecasting models based on lagged inputs. Their good performance in linear settings, particularly for models from the ARMA family, is not surprising given the way the relevance measures and the imputation strategies are constructed. However, the main contribution of these methods lies in their applicability to black-box models. Across the different neural network architectures considered in this work, the ARF and PARF were consistently able to identify the lags that were expected to be relevant in one-step-ahead forecasting tasks. In addition, the proposed methods are computationally simpler than standard SHAP-based approaches, which makes them especially attractive in practice.

The modifications introduced to adapt Shapley-based ideas to the time series setting also proved to be useful. In particular, the MSE-based contributions provided especially informative relevance patterns, notably in scenarios involving seasonality or nonlinear data-generating processes. With respect to the imputation methods, both approaches (model-based one-step-ahead shown in the main body of the paper, and the one based on a fitted multivariate-linear-model, which results are shown in the Appendix~A) produced similar results in simple linear examples, but the one-step-ahead prediction imputation showed clearer advantages in nonlinear settings and in cases involving more complex forecasting models. This finding suggests that model-based one-step-ahead imputation provides a natural and effective way to handle absent features in time series relevance analysis.

At the same time, this strategy relies on the forecasting model having sufficiently good predictive performance. If the fitted model produces poor forecasts, the imputed values may introduce noise and lead to misleading relevance estimates. For this reason, the experimental analysis in this work was based on carefully selected models and tuned hyperparameters, so that the resulting relevance measures would reflect meaningful predictive structures rather than deficiencies in model fit.

Overall, the proposed framework provides encouraging evidence that lag-based relevance measures can be extended beyond linear models and successfully applied to modern black-box predictors. These results open several directions for future work, including the study of alternative relevance formulations, the incorporation of exogenous predictors beyond lagged observations, and the extension of the methodology to multivariate time series settings.

\section*{Acknowledgements}

This work was supported by Agencia Estatal de Investigación and it is part of the project \textbf{PID2023-148158OB-I00} from \textit{CONTRATOS PREDOCTORALES PARA LA FORMACIÓN DE DOCTORES/DOCTORAS 2024 (FD 2024) del Programa Estatal para Impulsar la Investigación Científico-Técnica y su Transferencia, del Plan Estatal de Investigación Científica, Técnica y de Innovación 2021-2023.}

\begin{appendices}

\section{ARF's and PARF's with linear imputation}\label{secA1}

\subsection*{AR(3) Simulation}

\begin{figure}[H]
\centering
\includegraphics[width=0.8\linewidth]{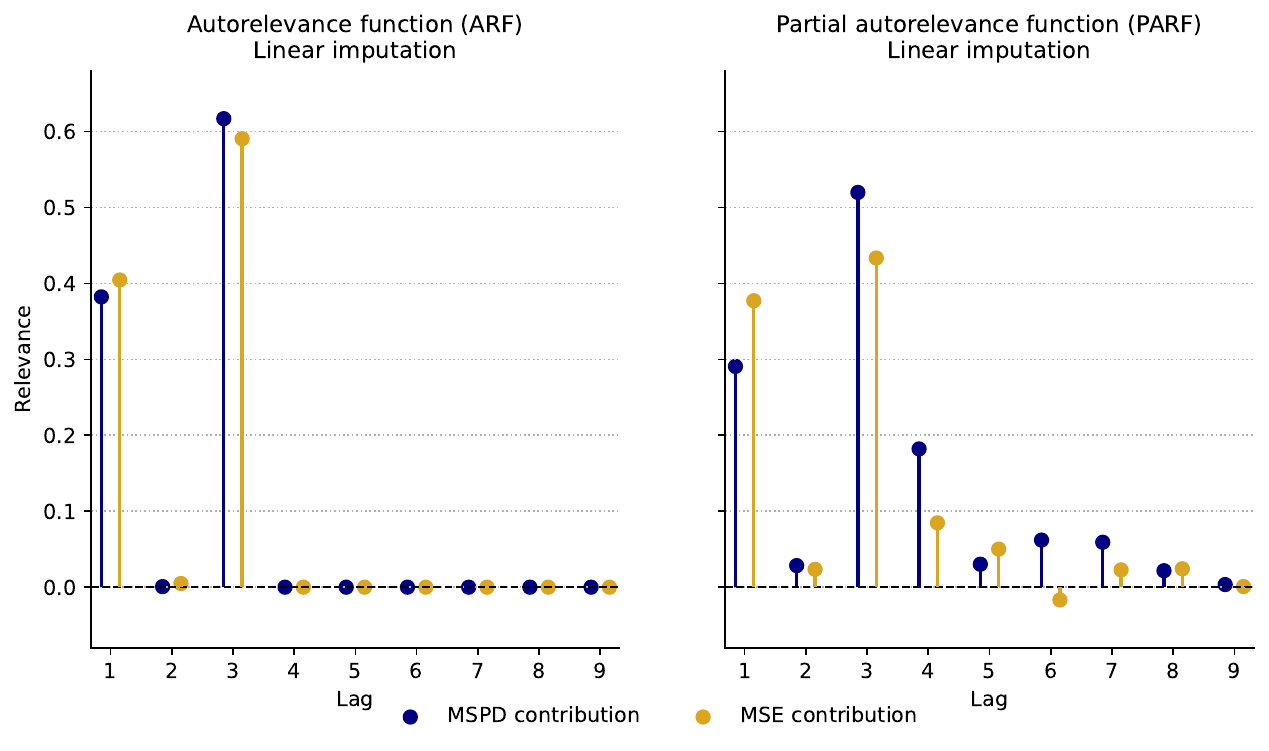}
\caption{Autorelevance and partial autorelevance functions - ARMA fitted model - AR$(3)$ series - Linear imputation}
\label{fig:doslin}
\end{figure}

\begin{figure}[H]
\centering
\includegraphics[width=0.8\linewidth]{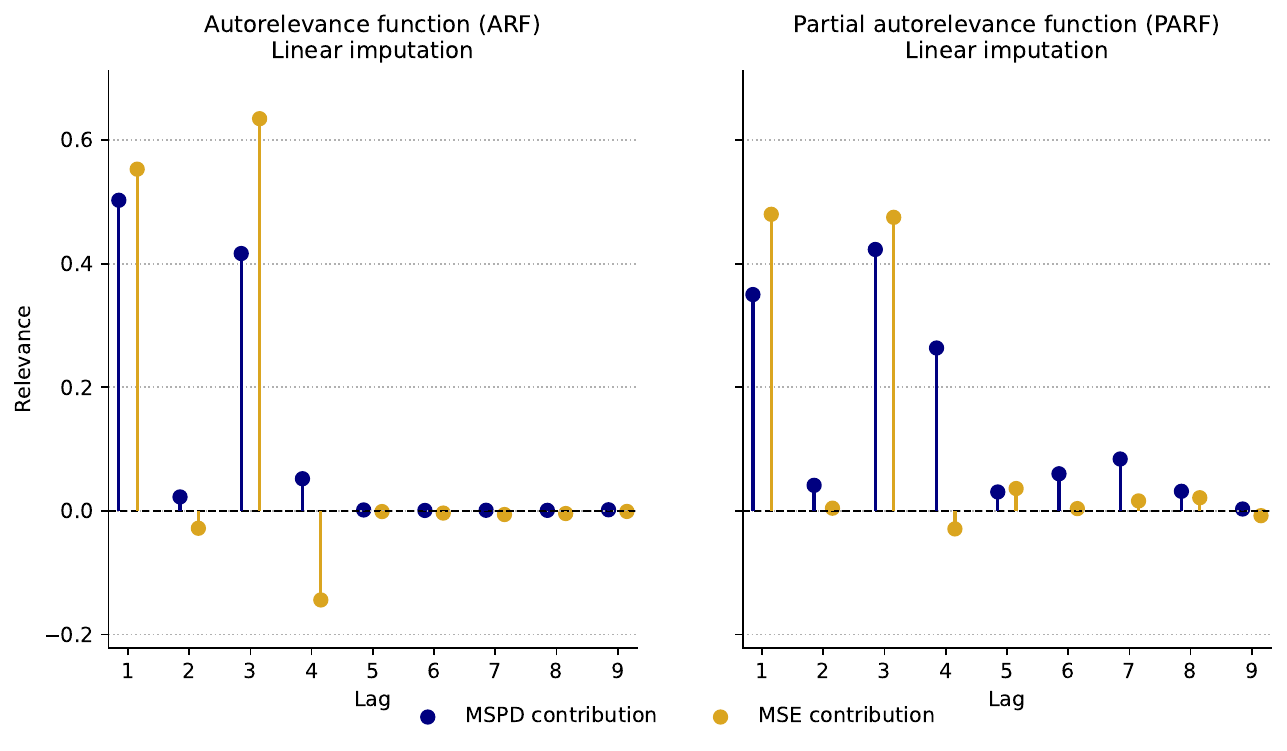}
\caption{Autorelevance and partial autorelevance functions - NN fitted model -AR$(3)$ series - Linear imputation}
\label{fig:cinclin}
\end{figure}

\subsection*{ARMA(3,3) Simulation}

\begin{figure}[H]
\centering
\includegraphics[width=0.8\linewidth]{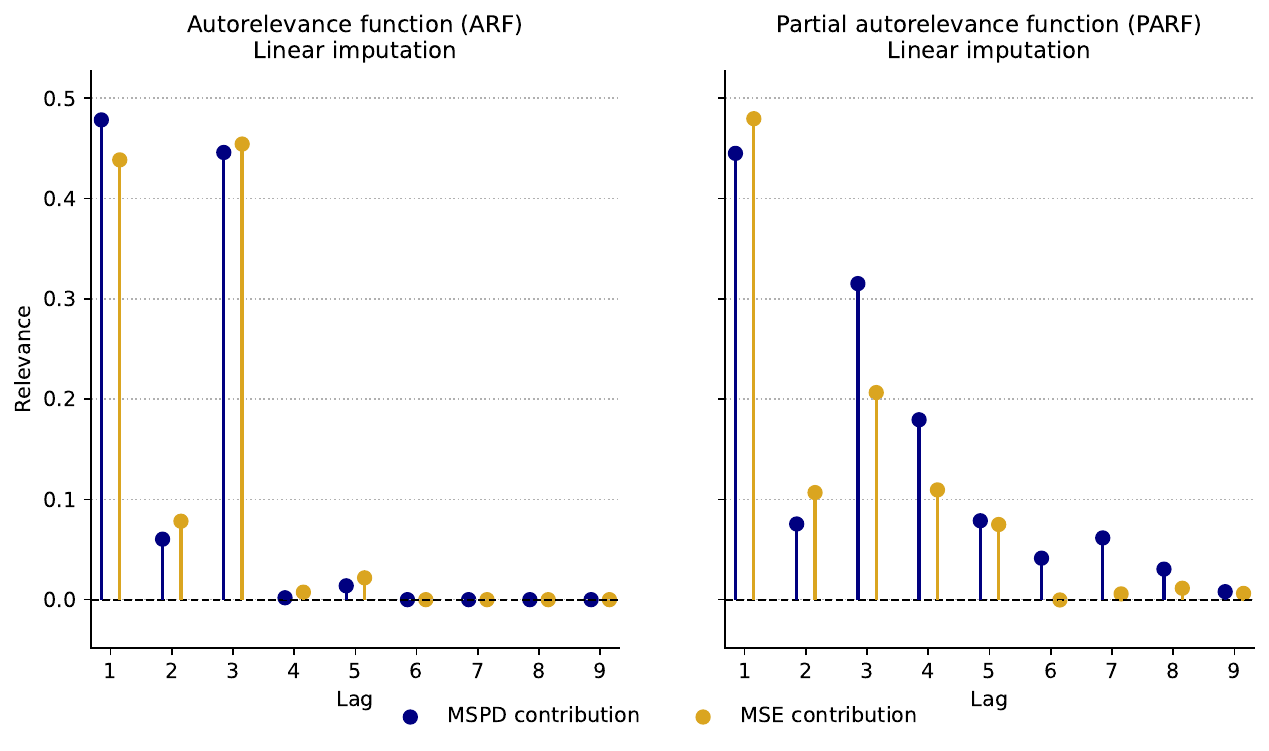}
\caption{Autorelevance and partial autorelevance functions - ARMA fitted model - ARMA$(3,3)$ series - Linear imputation}
\label{fig:linearex05AR}
\end{figure}

\begin{figure}[H]
\centering
\includegraphics[width=0.8\linewidth]{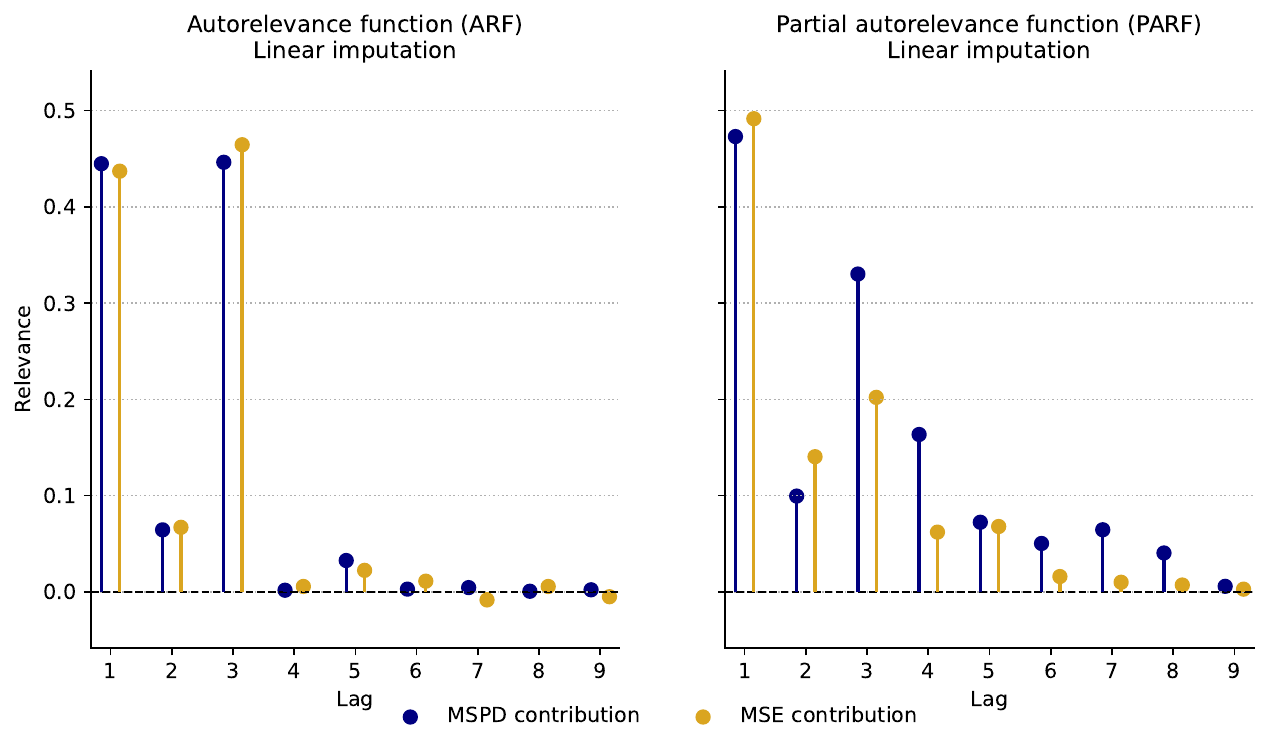}
\caption{Autorelevance and partial autorelevance functions - NN fitted model - ARMA$(3,3)$ series - Linear imputation}
\label{fig:linearex05NN}
\end{figure}

\subsection*{Non linear time series simulation}

\begin{figure}[H]
\centering
\includegraphics[width=0.8\linewidth]{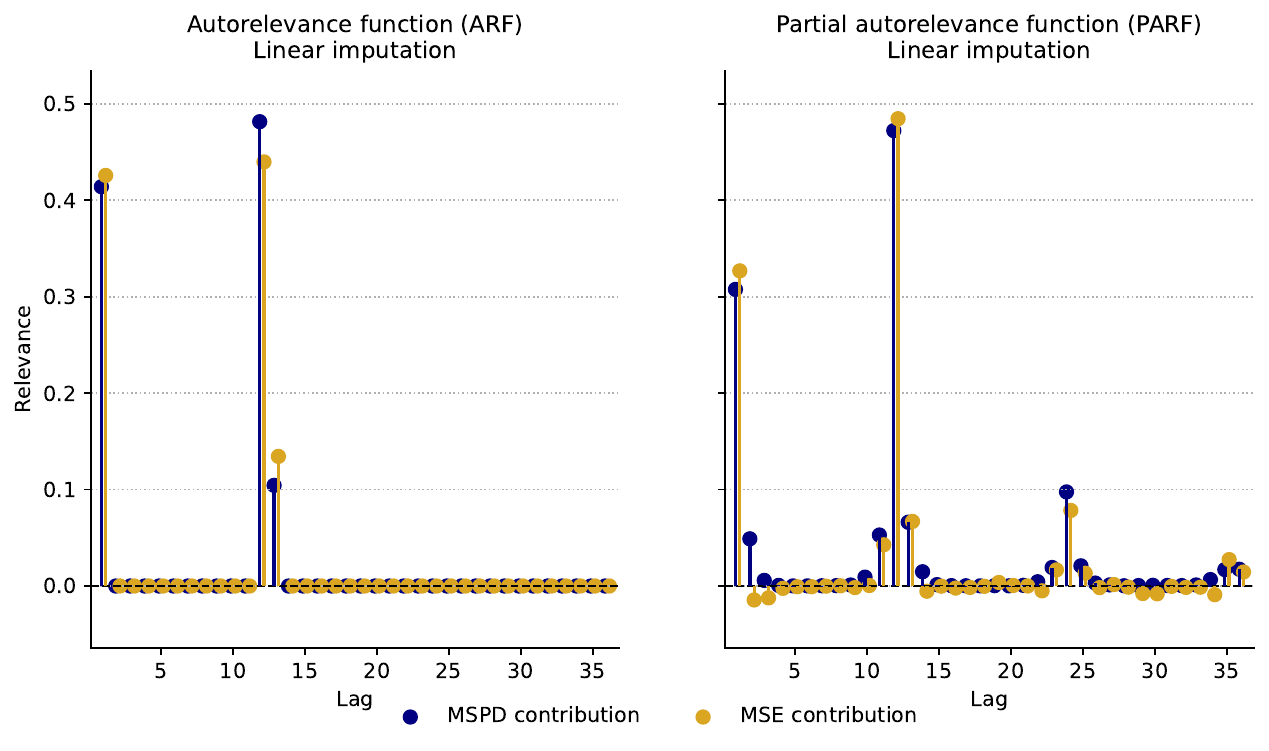}
\caption{Autorelevance and partial autorelevance functions - ARMA fitted model - Non linear series - Linear imputation}
\label{fig:linearex08AR}
\end{figure}

\begin{figure}[H]
\centering
\includegraphics[width=0.8\linewidth]{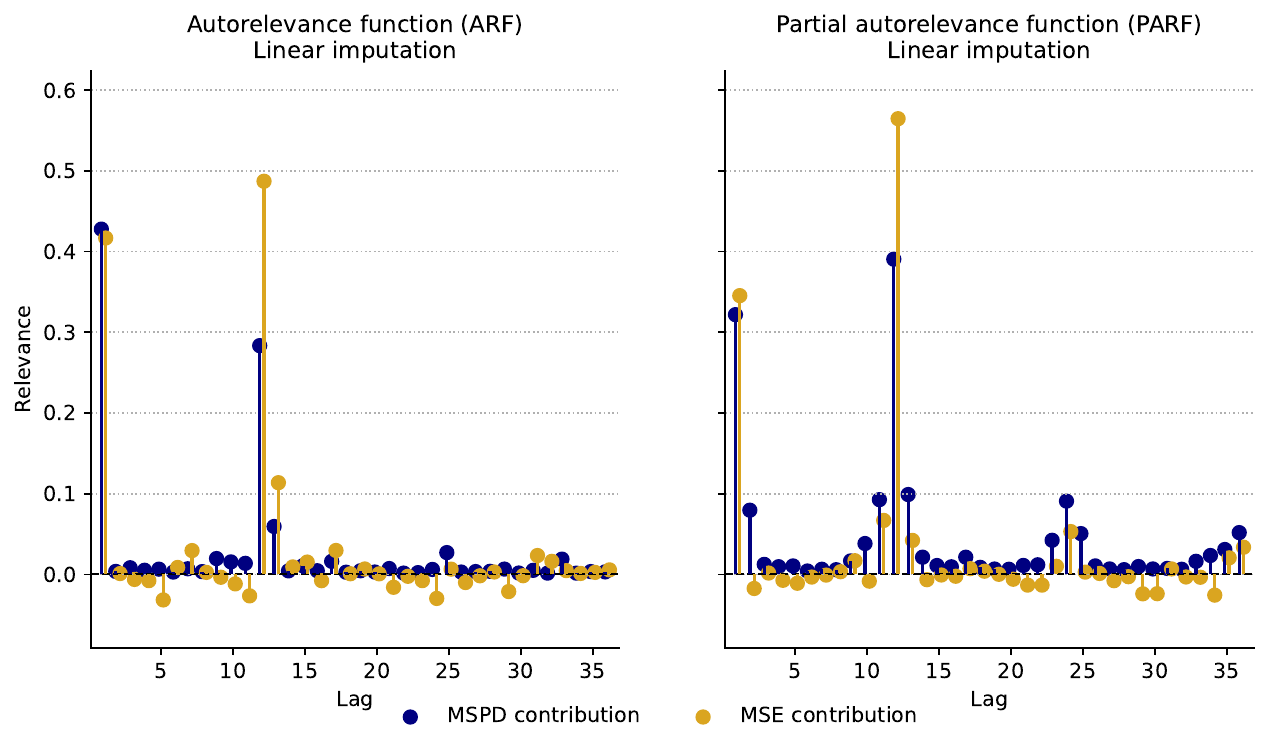}
\caption{Autorelevance and partial autorelevance functions - NN fitted model - Non linear series - Linear imputation}
\label{fig:linearex08AR}
\end{figure}

\subsection*{Airline passengers series}

\begin{figure}[H]
\centering
\includegraphics[width=0.8\linewidth]{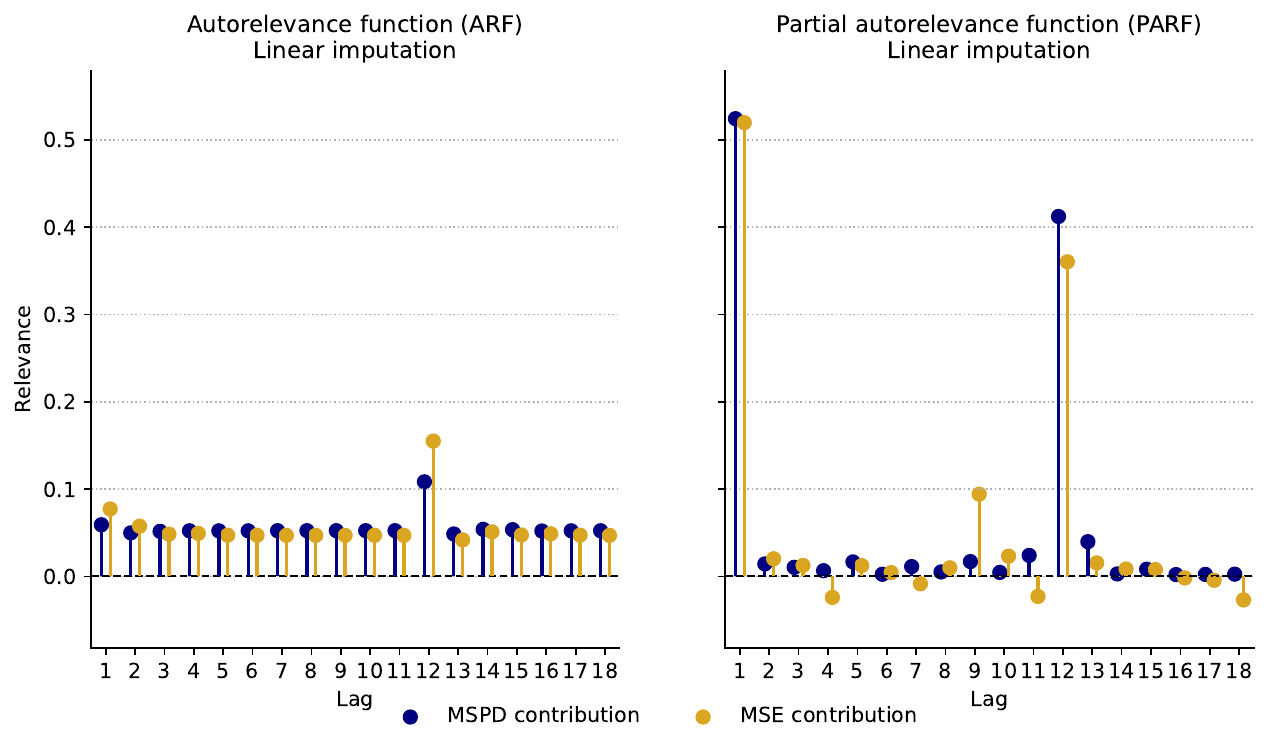}
\caption{Autorelevance and partial autorelevance functions - ARMA fitted model - Airline passengers series - Linear imputation}
\label{fig:airline1ARF}
\end{figure}

\begin{figure}[H]
\centering
\includegraphics[width=0.8\linewidth]{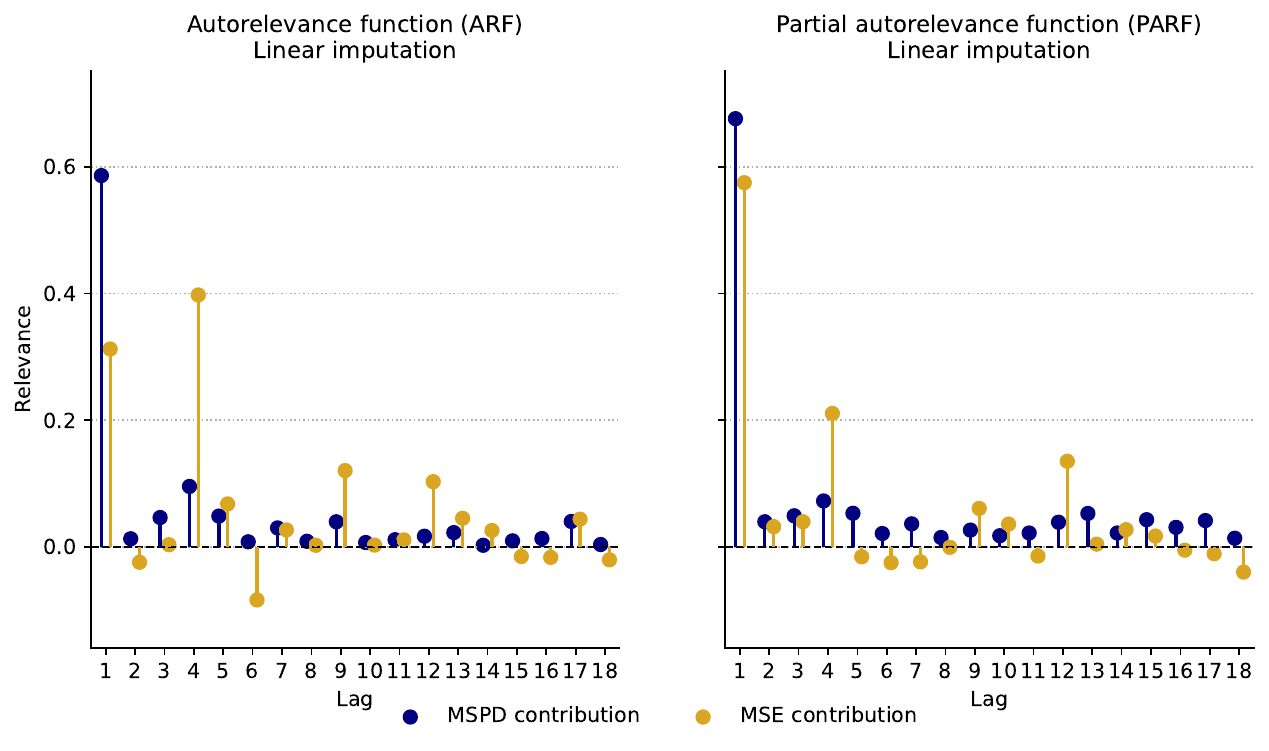}
\caption{Autorelevance and partial autorelevance functions - NN fitted model - Airline passengers series - Linear imputation}
\label{fig:airline2PARF}
\end{figure}

\newpage

\section{Details of fitted models}\label{secA2}

\subsection*{ARMA main parameters}

\begin{table}[h!]
\centering
\begin{adjustbox}{width=0.8\textwidth}
\small
\begin{tabular}{|c|l|l|l|}
\hline
\textbf{Example}                              & \multicolumn{1}{c|}{\textbf{Orders}} & \multicolumn{1}{c|}{\textbf{Seasonal orders}} & \multicolumn{1}{c|}{\textbf{MSE test set}} \\ \hline
\rowcolor[HTML]{FFFFFF} 
\textbf{AR(3)}                                & (3,0,0)                              &                                               & 0,644                                      \\ \hline
\rowcolor[HTML]{FFFFFF} 
\textbf{ARMA(3,3)}                            & (5,0,0)                     &                                             & 0,4705                                     \\ \hline
\textbf{Non-linear ts}               & (1, 0, 0)x(1, 0, 0)                              &    12                                           & 0,4699                                     \\ \hline
\textbf{Airlines ts}                           & (0, 0, 1)x(0, 0, 1)                             & 12                                     & 0,5719                                     \\ \hline
\end{tabular}
\end{adjustbox}
\end{table}

\subsection*{Neural networks hyperparameters}

\begin{table}[h!]
\centering
\begin{adjustbox}{width=0.8\textwidth}
\small
\begin{tabular}{|c|l|l|l|l|l|}
\hline
\textbf{Example}                              & \multicolumn{1}{c|}{\textbf{Input}} &  \multicolumn{1}{c|}{\textbf{Epoch}} & \multicolumn{1}{c|}{\textbf{LR}} & \multicolumn{1}{c|}{\textbf{Batch size}} & \multicolumn{1}{c|}{\textbf{MSE test set}} \\ \hline
\cellcolor[HTML]{FFFFFF}\textbf{AR(3)}        & \cellcolor[HTML]{FFFFFF}9           & \cellcolor[HTML]{FFFFFF} 150                                 & 0,0001                           & 16                                       & 0,7013                                     \\ \hline
\cellcolor[HTML]{FFFFFF}\textbf{ARMA(3,3)}    & \cellcolor[HTML]{FFFFFF}12           & \cellcolor[HTML]{FFFFFF} 150                                 & 0,0003                           & 16                                       & 0,4755                                     \\ \hline
\textbf{Non-linear ts}               & 48                                      & 200                                 & 0,0003                           & 16                                       & 0,4786                                     \\ \hline
\textbf{Airlines ts}               & 18                                      & 200                                 & 0,0003                           & 16                                       & 0,5100                                     \\ \hline
\end{tabular}
\end{adjustbox}
\end{table}

\begin{table}[h!]
\centering
\begin{adjustbox}{width=0.8\textwidth}
\small
\begin{tabular}{|c|l|}
\hline
\textbf{Example}                & \multicolumn{1}{c|}{\textbf{Details}}                                                  \\ \hline
\rowcolor[HTML]{FFFFFF} 
\textbf{AR(3)}                  & \makecell{Stacked LSTM 2 layers with 1 dense regularized layer}                                                                 \\ \hline
\rowcolor[HTML]{FFFFFF} 
\textbf{ARMA(3,3)}              & \makecell{Stacked LSTM 2 layers with 1 dense regularized layer}                                                                  \\ \hline
\textbf{Non-linear ts}   &  \makecell{Dilated Temporal Convolutional NN forecasting model that\\ learns temporal patterns from lagged univariate sequences through convolutional\\blocks with increasing dilation rates.} \\ \hline
\textbf{Airlines ts} &  \makecell{Dilated Temporal Convolutional NN forecasting model that\\ learns temporal patterns from lagged univariate sequences through convolutional\\blocks with increasing dilation rates.}               \\ \hline
\end{tabular}
\end{adjustbox}
\end{table}




\end{appendices}

\newpage

\bibliography{sn-bibliography}

\end{document}